\documentclass[11pt]{article}

\usepackage[preprint]{acl}

\usepackage{times}
\usepackage{latexsym}

\usepackage[T1]{fontenc}
\usepackage[utf8]{inputenc}

\usepackage{microtype}

\usepackage{inconsolata}

\usepackage{graphicx}

\usepackage{multirow}
\usepackage{bm}
\usepackage{booktabs}
\usepackage{makecell}
\usepackage{adjustbox}
\usepackage{algorithm}
\usepackage{algorithmic}

\usepackage{caption}

\usepackage{tcolorbox}
\tcbuselibrary{listingsutf8}
\tcbuselibrary{skins,breakable}

\newcolumntype{C}[1]{>{\centering\arraybackslash}p{#1}}

\newenvironment{promptbox}{%
    \tcblisting{
        enhanced,
        listing only,
        colback=gray!10,
        colframe=black,
        top=2mm,
        bottom=2mm,
        boxrule=0.5pt,
        before skip=10pt,
        after skip=10pt,
        after={\par\vspace{0.5\baselineskip}\noindent},
    }
}
{\endtcblisting}

\title{Difficulty-Controllable Cloze Question Distractor Generation}

\author{
\textbf{Seokhoon Kang\textsuperscript{1}},
\textbf{Yejin Jeon\textsuperscript{3,4}\setcounter{footnote}{1}\thanks{This work was conducted while at POSTECH.}},
\textbf{Seonjeong Hwang\textsuperscript{1}},
\textbf{Gary Geunbae Lee\textsuperscript{1,2}}
\\
\\
\textsuperscript{1}Graduate School of Artificial Intelligence, POSTECH, South Korea\\
\textsuperscript{2}Department of Computer Science and Engineering, POSTECH, South Korea\\
\textsuperscript{3}Mila Quebec AI Institute, Canada\\
\textsuperscript{4}McGill University, Canada\\
\texttt{\{sh.kang, seonjeongh, gblee\}@postech.ac.kr}\\
\texttt{yejin.jeon@mila.quebec}
}

\begin{document}
\maketitle
\begin{abstract}
Multiple-choice cloze questions are commonly used to assess linguistic proficiency and comprehension. However, generating high-quality distractors remains challenging, as existing methods often lack adaptability and control over difficulty levels, and the absence of difficulty-annotated datasets further hinders progress. To address these issues, we propose a novel framework for generating distractors with controllable difficulty by leveraging both data augmentation and a multitask learning strategy. First, to create a high-quality, difficulty-annotated dataset, we introduce a two-way distractor generation process to produce diverse and plausible distractors. These candidates are filtered and then categorized by difficulty using an ensemble QA system. Second, this newly created dataset is used to train a difficulty-controllable generation model via multitask learning. Experimental results demonstrate that our method generates high-quality distractors across difficulty levels and substantially outperforms GPT-4o in aligning distractor difficulty with human perception.
\end{abstract}

\section{Introduction}

The widespread adoption of e-learning platforms has transformed traditional methods used in education, enabling access to knowledge and breaking down geographical and temporal barriers~\cite{desouza-2021-exploration}. Moreover, with the growing number of learners, scalable and effective assessment methods have become increasingly critical.  As such, cloze questions are being widely used in language proficiency evaluation for their ability to assess diverse linguistic skills. Specifically, this particular assessment format involves the removal of specific words from a passage and requires learners to fill in contextually appropriate terms~\cite{taylor-1953-cloze}. Among their variants, multiple-choice cloze questions are particularly favored for their ease of scoring and enhanced objectivity in large-scale testing.

A key challenge in constructing such questions lies in generating plausible distractors, or incorrect options~\cite{haladyna-2004-developing} that challenge students to engage with the material and enhance reading comprehension by distinguishing the correct answer from similar but incorrect choices.
Distractors that are neither too obvious nor overly misleading are not only essential for maintaining assessment validity, but also serve as a critical determinant of overall question difficulty~\cite{rezigalla-2024-item, susanti-2017-controlling, yuni-2016-item}. 
However, manually crafting high-quality distractors is time-consuming and resource-intensive, which makes it impractical for large-scale deployment.

Towards this, several studies have proposed automated distractor generation systems~\cite{yeung-2019-difficulty, siyu-2021-knowledge, chiang-2022-cdgp, wang-2023-distractor}. 
While effective at replicating distractors in the training data~\cite{siyu-2021-knowledge, chiang-2022-cdgp, wang-2023-distractor}, these approaches struggle to generate distractors with varying difficulty levels, which limits their use in personalized learning environments.
Although \citet{yeung-2019-difficulty} explored difficulty-aware distractor generation, they rely on predefined candidate lists which limits diversity and difficulty range.

The challenge pertaining to difficulty control arises from multiple factors. 
First, difficulty is inherently subjective and lacks a universally accepted metric, which complicates its operationalization in automated systems~\cite{alkhuzaey-2024-text}. 
Second, the limited number of distractors per question in the dataset restricts the variety of examples available for training, which hinders the model’s ability to generalize across a spectrum of difficulty levels.

To address these challenges, we propose a framework for difficulty-controllable distractor generation that combines data augmentation and multitask learning.
Since our framework targets language proficiency assessment, we define difficulty based on the semantic plausibility of the distractor within the context, which is consistent with the existing benchmark~\cite{xie-2018-large}. This approach allows for more objective control, independent of individual learner differences such as vocabulary size, thereby mitigating the subjectivity commonly associated with difficulty modeling.
Additionally, since the most commonly used datasets for cloze questions focus on word-level completions, and modifying a single word to achieve a continuous spectrum of difficulty poses inherent challenges, we adopt a binary classification approach for measuring distractor difficulty.

To construct a difficulty-annotated dataset, we introduce a two-way distractor generation method that incorporates an information restriction strategy to produce diverse and contextually plausible distractors. Candidates are refined through a filtering stage to ensure semantic and grammatical quality, and then clustered by difficulty using an ensemble of QA models with score normalization for improved accuracy.
We then train a generation model on the augmented dataset using a multitask learning objective, enabling it to not only produce distractors at specified difficulty levels but also to distinguish between correct answers, easy, and hard distractors.

% 변경 문단 (내용 추가)
Quantitative and qualitative evaluation results demonstrate the effectiveness of our approach. 
Our augmentation method expands the CLOTH dataset~\cite{xie-2018-large} to an average of 12 distractors per question, and is categorized by difficulty. We also observe that our high-attention information restriction strategy allows for flexible control over both distractor difficulty and semantic coverage.
Our method significantly outperforms GPT-4o, with 73.25\% and 64.23\% difficulty accuracy in generating hard and easy distractors, respectively. Our model also significantly reduces the invalid distractor ratio, with no invalid distractors in easy distractors and 1.6\% for hard distractors.

Our contributions are summarized as follows:
\begin{itemize}
    \setlength\itemsep{0em}
    \item We introduce a novel two-way data augmentation pipeline that enables flexible difficulty control via information restriction.
    \item We design a multitask learning strategy that allows the model to distinguish answers from distractors and assess their relative difficulty.
    \item Through comprehensive automatic and human evaluations, we demonstrate that our proposed method significantly outperforms GPT-4o in aligning distractor difficulty while maintaining a low invalid distractor ratio.
    \item We release both the augmented dataset and the trained model to support future research.\footnote{\scriptsize \url{https://github.com/ksh108405/DCDG}}
\end{itemize}

\section{Related Work}

\begin{figure*}[t]
  \centering
  \includegraphics[width=0.9\linewidth]{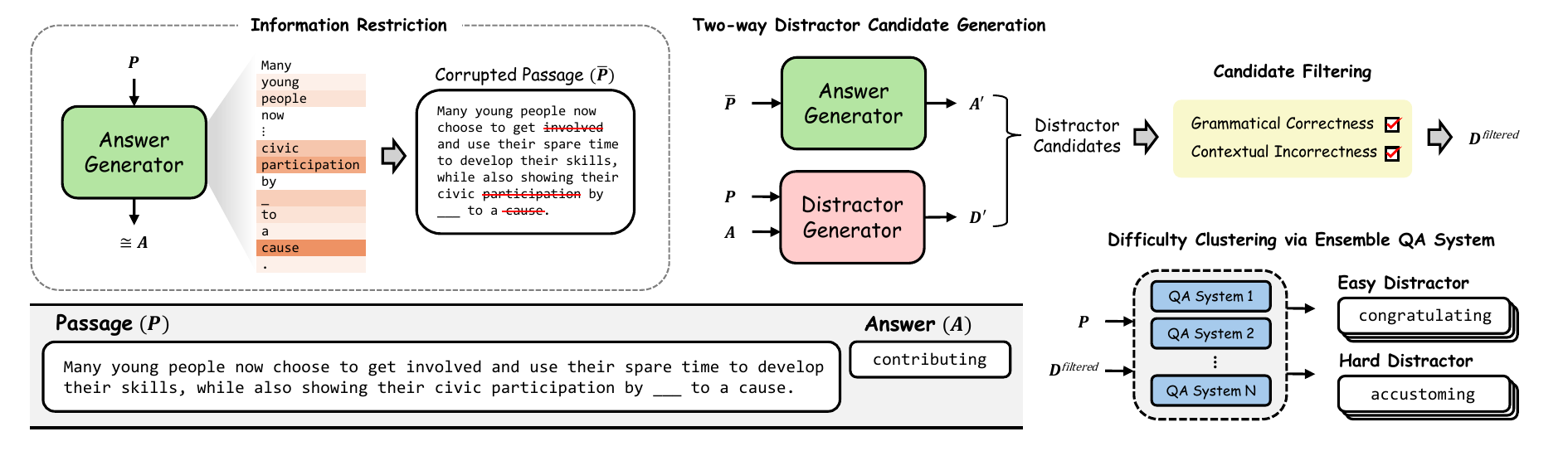}
  \caption {Overview of the dataset augmentation pipeline.}
  \label{fig:dataset_arch}
  \vspace{-7pt} 
\end{figure*}

% \subsection{Cloze Question Distractor Generation}

Early studies used linguistic heuristics and domain-specific resources for distractor generation~\cite{pino-2009-semi, correia-2010-automatic}. In particular, they leveraged thesauri, taxonomies, or predefined vocabularies to select distractors that were semantically related to the correct answer. However, the reliance on domain-specific resources limited their applicability across diverse subjects. To address the limitations of such heuristic methods, subsequent studies incorporated general-purpose knowledge bases. For instance, \citet{siyu-2021-knowledge} proposed a framework that leverages knowledge bases like Probase~\cite{wu-2018-probase} and WordNet~\cite{fellbaum-1998-wordnet} to generate candidate distractors. Their approach used a learning-to-rank model to select distractors and demonstrated improved performance across multiple domains. Despite these advancements, the dependency on structured knowledge bases still constrains the model's adaptability in domains where such resources are sparse or outdated.

Recent literature has shifted towards utilizing transformer-based pretrained language models (PLMs) for distractor generation. For instance, \citet{chiang-2022-cdgp} used PLMs to generate distractors, and \citet{wang-2023-distractor} formulated cloze distractor generation as a Text2Text task and used PLMs enhanced with pseudo Kullback-Leibler divergence to prevent generating distractors that were too similar to each other. Nevertheless, these methods still struggle with difficulty control, leading to limited adaptability across varying educational needs.
% Their approach also incorporated candidate augmentation and multitask training, leading to significant performance gains. 

As such, difficulty-controllable question generation has emerged as a relatively new research domain. With the growing popularity of PLMs, recent works such as ~\citet{uto-2023-difficulty}, ~\citet{tomikawa-2024-difficulty}, and~\citet{park-2024-large} combine PLMs with Item Response Theory (IRT) to model and control question difficulty.
However, the correlation between human and PLM-simulated IRT parameters remains unverified in the cloze domain, and validating this is challenging since reliable IRT estimation necessitates large-scale human response data~\cite{sireci-1992-utility}, which is unavailable in public datasets. Furthermore, such methods incur high computational costs, often requiring hundreds of models to simulate student responses~\cite{tomikawa-2024-difficulty}. Given these constraints, we adopt a more resource-efficient approach using a discrete difficulty metric, which is calibrated with expert annotations.

\section{Methodologies}

\subsection{Dataset Augmentation}
\label{sec:dataset_augmentation}

To augment distractors for cloze questions, we propose a three-step system designed to 1) generate, 2) filter, and 3) cluster distractors by difficulty. Each step is detailed in its respective subsections. This augmentation process is illustrated in Figure~\ref{fig:dataset_arch}.

\paragraph{Two-Way Candidate Generation}
In order to generate diverse and high-quality candidates for distractor augmentation, we adopt a two-way approach that integrates two distinct methods. First, we use a \textit{distractor generator}, which is a language model fine-tuned on the distractors present in the original dataset. Yet, as this generator is specifically trained to replicate the characteristics of ground-truth distractors, its output is constrained by the semantic and difficulty distribution of the training data, which limits the generator's ability to produce diverse distractors for novel contexts or difficulty levels.
Therefore, to address this limitation, we propose an information-restriction generation technique. This method re-purposes an \textit{answer generator}, which was originally trained to generate correct answers, by strategically deleting its available information to produce distractors. This approach is built on two established findings; distractors deviate from the correct answer by contradicting specific parts of the passage~\cite{ondov-2024-pedagogically}, and there exists a correlation in reading patterns between humans and transformer-based models~\cite{zou-2023-human, bensemann-2022-eye}.

Building on this insight, we repurposed the answer generator in a two-stage process to generate distractors. 
%In the first stage, the generator processes the entire passage to extract attention scores for each token in order to identify the most crucial words. 
In the first stage, it processes the entire passage and generates the answer autoregressively. During this process, attention scores are computed at each generation step by summing across all layers and heads, and then accumulated across all generated answer tokens to derive the final attention scores for identifying crucial passage words.
In the second stage, the passage is selectively pruned by removing words with the highest attention scores until a predefined deletion ratio is met. The pruned passage is then fed back into the answer generator to generate distractors that intentionally conflict with the passage. By adjusting the deletion ratio, we control the degree of inconsistency with the passage, thus producing distractors of varying difficulty.
Pseudo-code for this information restriction is provided in Appendix~\ref{sec:info_delete_code}.

\paragraph{Candidate Filtering}
An essential property of effective distractors is that they must not align with the passage more closely than the correct answer. 
To achieve this, we use a two-step candidate filtering process that verifies whether each distractor is grammatically appropriate for the blank and ensures it cannot be a valid answer.

First, grammatically incorrect candidate distractors are eliminated using the rule-based LanguageTool~\cite{naber-2003-rule} grammatical error corrector. Second, GPT-4o mini is used to assess the semantic plausibility of the remaining candidates by prompting the model to judge whether a distractor could serve as a valid answer in the given cloze passage. Those considered plausible answers are removed so as to retain only contextually inappropriate distractors. This two-step filtering process thus yields distractors that are both well-formed and clearly unsuitable as correct answers.

\paragraph{Difficulty Clustering}
To measure distractor difficulty, we use an ensemble of PLMs as done in prior studies~\cite{yeung-2019-difficulty, chiang-2022-cdgp, cavusoglu-2024-disgem}.
Each PLM in the system is fine-tuned with a multiple-choice cloze QA setup to assign a score to each option, which represents its likelihood of being selected as the correct answer. To cluster distractors by difficulty, we divide the entire pool into three equal-sized subsets based on their scores. The hard distractor set is comprised of distractors from the top third subset, while the easy distractor set is derived from the bottom third subset. To maximize diversity, we computed STS using cosine similarity on sentence embeddings from the \texttt{all-mpnet-base-v2} model in the Sentence-Transformer~\cite{reimers-2019-sentence}. Specifically, we calculated a pairwise cosine similarity matrix for all distractor candidates and filtered out any option with a similarity score exceeding 0.8 with any other selected candidate. Distractors in the middle range are excluded from use, as they do not clearly align with either difficulty category.

A key aspect of the ensemble system is the normalization of confidence scores from each model, which is critical for ensuring balance across models. Given that the confidence scores for distractors exhibit a highly right-skewed distribution, we apply Box-Cox transformation~\cite{box-1964-analysis}. This transformation is applied independently for each passage and model, with parameter $\lambda$ optimized via maximum likelihood estimation to best fit a Gaussian distribution. For details, see Appendix~\ref{sec:normalization}.

\subsection{Training Strategy}

\begin{figure}[t]
  \centering
  \includegraphics[width=\linewidth]{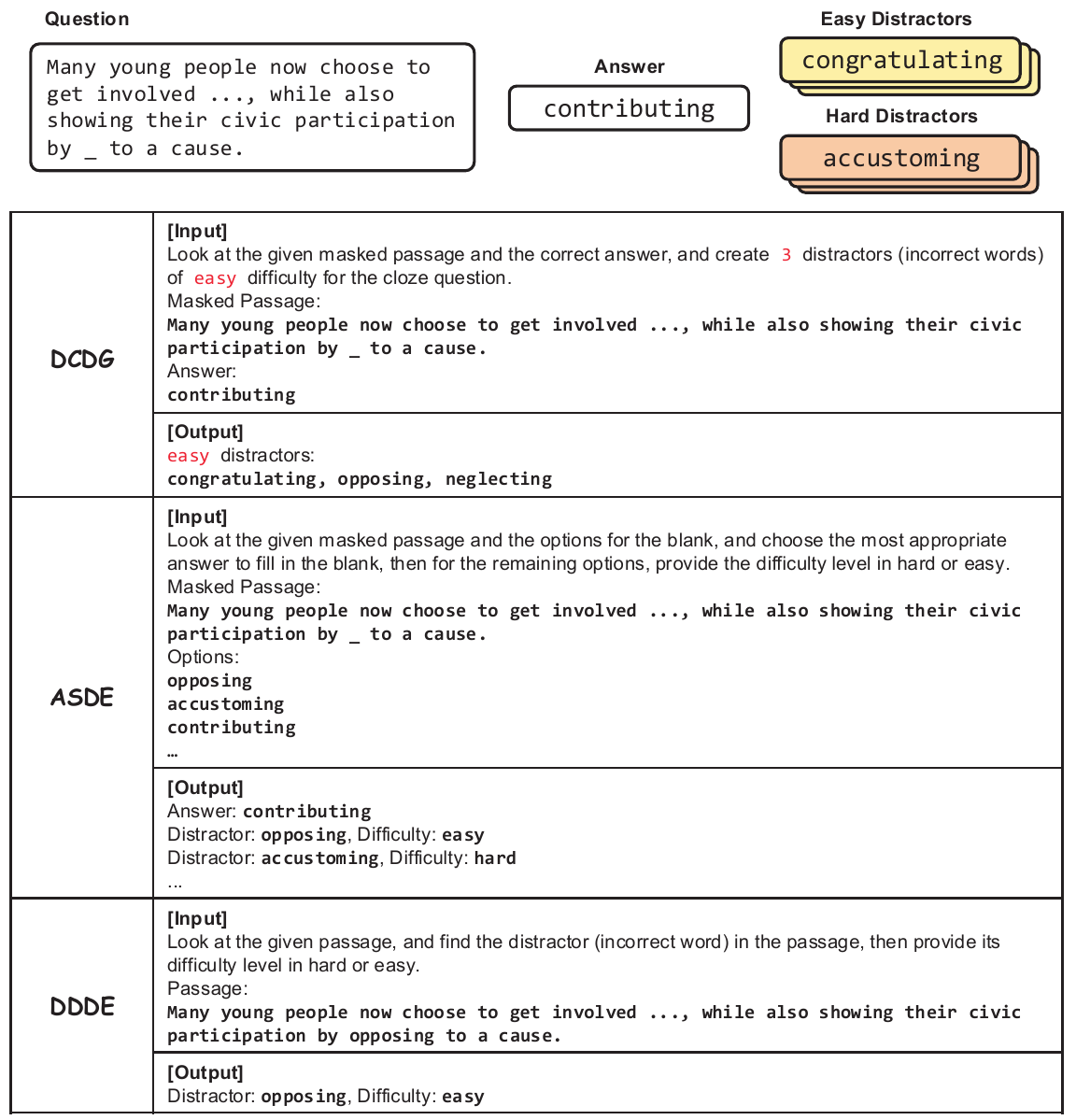}
  \caption {Overview of training methods of difficulty-controllable cloze distractor generation model. Template variables are highlighted in red.}
  \label{fig:model_train}
  \vspace{-10pt}
\end{figure}

To effectively train our difficulty-controllable distractor generation model using the augmented dataset and to enhance its understanding of distractor difficulty, we propose a novel multitask training strategy. This strategy consists of one main task and two auxiliary tasks, all formulated in a sequence-to-sequence (seq2seq) framework. Input-output formats for each task are shown in Figure~\ref{fig:model_train}.

\paragraph{Main Task}
At the core of our method is difficulty controllable distractor generation (DCDG), where an LLM is fine-tuned on the previously augmented dataset (\S\ref{sec:dataset_augmentation}).
For each passage, the model is trained separately on the entire set of distractors from both the hard and easy subsets, treating each difficulty level as a distinct training instance.
Specifically, given the cloze passage, the number of distractors to generate, the desired difficulty level, and the ground-truth answer, the model learns to generate appropriate distractors that align with the specified difficulty.

\paragraph{Auxiliary Tasks}
To enhance the model's understanding of distractor semantics and difficulty, we introduce two auxiliary tasks inspired by~\citet{wang-2023-distractor}, and extend their formulation to explicitly incorporate difficulty signals.
The primary goal of this multitask setup is to align the difficulty control tokens with the semantic plausibility of distractors within the given context, to differentiate between correct answers and distractors while recognizing relative difficulty.
In the first auxiliary \textbf{answer selection and distractor difficulty estimation (ASDE)} task, the model learns to identify the correct answer and simultaneously assess the difficulty of given distractors.
It receives a cloze passage with shuffled choices containing the ground-truth answer and distractors of varying difficulty.
The model is trained to 1) identify the correct answer and 2) label each distractor with an estimated difficulty.

Unlike ASDE, in the \textbf{distractor detection and difficulty estimation (DDDE)} task, the model is given a cloze passage with a distractor word inserted into the blank.
This task trains the model to 1) detect the distractor and 2) estimate its difficulty level, thereby aligning its predictions with the difficulty annotated in the augmented dataset.
Note that the full dataset is used to train the main task, while the auxiliary tasks are trained on their respective converted subsets. 
All tasks are optimized using a standard cross-entropy loss, calculated within the unified seq2seq formulation.

\section{Experiments}

Experiments are conducted on the CLOTH dataset~\cite{xie-2018-large}, which is available for non-commercial research purposes only. This dataset consists of 7,131 unmasked passages and 99,433 cloze questions that are derived from middle and high school English exams in China, and covers a diverse range of topics and difficulty levels. Each passage contains multiple blanks for completion and is split into 5,513 passages for training, 805 for validation, and 813 for testing.

\paragraph{Two-Way Candidate Generation}
To augment the CLOTH dataset, we train Gemma 2 9B~\cite{gemmateam-2024-gemma2} for both the distractor and answer generators. To prevent data leakage, we applied 5-fold cross-validation. For each iteration, the model was trained on four folds and performed inference exclusively on the remaining held-out fold. This cycle was repeated five times to augment the entire dataset without any overlap between training and generation data. To enhance the robustness of the answer generator during information restriction, 50\% of the passages were randomly modified during training by removing words at varying rates between 0\% and 100\%. We used deletion ratios of [0.1, 0.2, 0.4] to create distractors of varying difficulty. Additionally, words surrounding the blank were preserved to avoid generating trivially easy distractors. Training was done with two NVIDIA A100-80GB GPUs, with a global batch size of 16 and a learning rate of 5e-6.

\paragraph{Candidate Filtering}

To ensure the quality of the augmented distractors, we applied a two-step filtering process. For grammatical validation, we used LanguageTool and deleted 2\% of the candidates which were ungrammatical or nonsensical. To assess contextual correctness, we prompted GPT-4o mini in a 2-shot cloze QA format to detect distractors that fit the blank better than the original answer. This filtering was repeated three times, resulting in the removal of 21.31\% of generated distractors.
Prompt templates are detailed in Appendix~\ref{sec:filtering}.

\paragraph{Difficulty Clustering}

\begin{figure}[t]
\centering
  \includegraphics[width=0.8\linewidth]{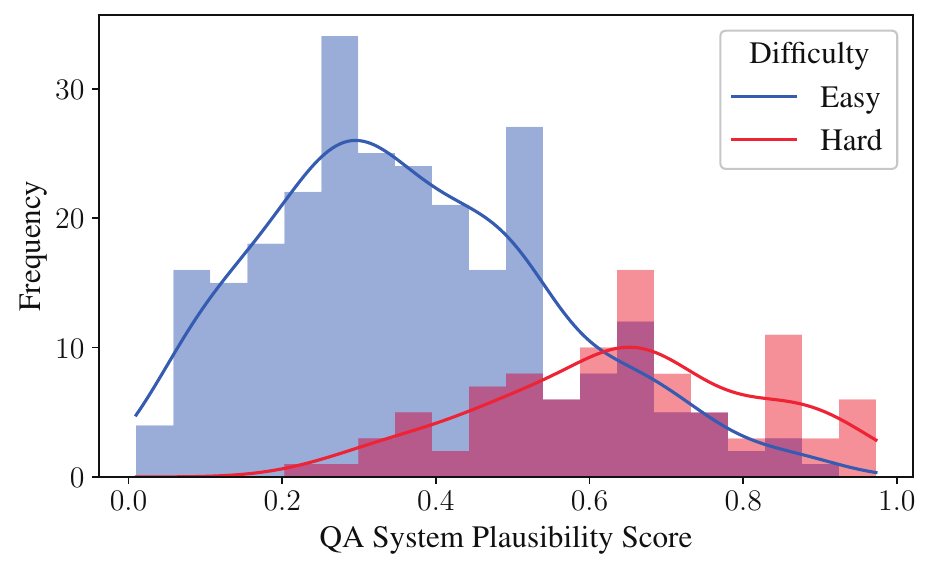}
  \caption {QA system annotation scores across two difficulty levels. The line represents the continuous distribution of the data using KDE.}
  \label{fig:mcqa_small_eval}
\end{figure}

To determine the difficulty levels of the distractors in the augmented CLOTH dataset, we used a diverse set of small PLMs across 11 model families. A total of 18 models were trained and evaluated.
To select the best-performing models which aligns with target users, a highly proficient ESL expert with over 20 years of academic English experience annotated the calibration set. Specifically, the expert labeled 359 distractors from 10 randomly sampled questions in the augmented validation set, categorizing them into two difficulty levels (easy and hard) based on perceived difficulty for ESL learners.
Each model was then evaluated using this labeled test set. For the evaluation metric, we measure the degree of separation between the two difficulty distributions by computing the best balanced accuracy at the optimal threshold that maximally distinguishes them.

Among the top-performing model combinations, we selected those with a standard deviation below 0.5\% for stability. This led to the selection of six models, including albert-xlarge-v2, albert-xxlarge-v2, conv-bert-base, electra-large-dis, roberta-large, and xlnet-large. Similarly, a 5-fold training approach was applied to prevent data leakage.
For specific training details, refer to Appendix~\ref{sec:qa_system}.

Figure~\ref{fig:mcqa_small_eval} presents the score distribution of the QA ensemble system for human-labeled distractors. The distribution of easy distractors is left-skewed relative to that of hard distractors, indicating that they tend to receive lower scores. This suggests that the system effectively identifies easy distractors as less plausible, demonstrating its ability to capture and reflect varying difficulty levels in alignment with human-labeled classifications.

\paragraph{Difficulty-controllable Distractor Generation}
Gemma 2 9B is trained on the augmented data for difficulty-controllable distractor generation using a single NVIDIA L40S GPU with a global batch size of 16. The learning rate was set to 5e-5 for DDDE multitask training and 3e-5 for the others. We also used early stopping to ensure proper evaluation across training objectives. To enhance efficiency, we applied low-rank adaptation~\cite{hu-2022-lora} with $r=16$ and $\alpha=16$ and a 0.1 warm-up ratio for efficiency and stability.

\section{Results and Analysis}

\subsection{Dataset Augmentation}
\label{sec:augment}

\begin{table}[t]
\begin{adjustbox}{width=0.8\columnwidth,center}
\begin{tabular}{lcccc}
\toprule
\multirowcell{2}[-3pt][l]{\textbf{Dataset}}                &          & \multirowcell{2}[-3pt]{\textbf{Original}} & \multicolumn{2}{c}{\textbf{Augmented}}                                \\ \cmidrule{4-5}
                                                           &          &                                           & \multicolumn{1}{c}{\textbf{Easy}} & \multicolumn{1}{c}{\textbf{Hard}} \\ \midrule
\multirowcell{2}[-3pt][l]{\# of Distractors\\per Question} & $\mu$    & 2.998                                     & 12.06                             & 12.02                             \\ \cmidrule{2-5}
                                                           & $\sigma$ & 0.063                                     & 3.758                             & 3.770                             \\ \midrule
\multirowcell{2}[-3pt][l]{Similarity\\with Answer}         & $\mu$    & 0.3504                                    & 0.2901                            & 0.3592                            \\ \cmidrule{2-5}
                                                           & $\sigma$ & 0.1006                                    & 0.0600                            & 0.0756                            \\
\bottomrule
\end{tabular}
\end{adjustbox}
\caption {Statistics of the augmented dataset. $\mu$ and $\sigma$ denotes the mean and standard deviation, respectively.}
\label{tab:data_dis_stat}
\end{table}

\paragraph{Dataset Statistics}
Table~\ref{tab:data_dis_stat} demonstrates the distinct improvements over the original CLOTH dataset in both distractor quantity and quality. The augmented dataset contains a greater number of distractors per question, and thus provides a more comprehensive set. Additionally, STS analysis with the \texttt{all-mpnet-base-v2} shows that easy distractors in the augmented dataset are less similar to the correct answers compared to those in the original dataset, whereas hard distractors exhibit greater similarity. Moreover, lower standard deviation of similarity scores across both difficulty levels suggests a more consistent control over the range of distractor difficulty.

\paragraph{Difficulty and Quality Evaluation}
We used GPT-4o~\cite{openai-2024-gpt4} to further evaluate the quality and difficulty of augmented distractors. From the CLOTH test set, we sampled 1,000 questions and presented GPT-4o with a cloze passage alongside four options, which are the ground-truth answer, one randomly selected original distractor, and one easy and hard distractor from the augmented dataset. To assess relative difficulty, GPT-4o ranked the options based on their fit in the blank. Additionally, we measured the invalid ratio by checking whether any option was equally or more appropriate than the ground-truth answer. All evaluations were conducted in a 1-shot setting for consistency. The evaluation prompts are provided in Appendix~\ref{sec:gpt_eval}.

\begin{table}[t]
\begin{adjustbox}{width=0.85\columnwidth,center}
\begin{tabular}{llccc}
\toprule
\multicolumn{2}{l}{\multirowcell{2}[-3pt][l]{\textbf{Dataset}}} & \multirowcell{2}[-3pt]{\textbf{Original}} & \multicolumn{2}{c}{\textbf{Augmented}} \\ \cmidrule{4-5}
                                                &               &                                           & \textbf{Easy}    & \textbf{Hard}       \\ \midrule
\multirowcell{3}[-5pt][l]{Relative\\Difficulty} & Hardest       & 26.53\%                                   & 3.42\%           & \textbf{70.05\%}    \\ \cmidrule{2-5}
                                                & Middle        & \textbf{52.56\%}                          & 23.32\%          & 24.12\%             \\ \cmidrule{2-5}
                                                & Easiest       & 21.21\%                                   & \textbf{73.17\%} & 5.63\%              \\ \midrule
\multicolumn{2}{l}{Invalid Ratio}                               & 0.9\%                                     & \textbf{0.0\%}   & 4.2\%               \\
\bottomrule
\end{tabular}
\end{adjustbox}
\caption {Relative difficulty and invalid distractor ratios for ground-truth and augmented distractors.}
\label{tab:data_dis_gpt4o}
\end{table}

Table~\ref{tab:data_dis_gpt4o} reveals clear differences in difficulty across distractor types. Augmented hard distractors were categorized as the \texttt{Hardest} 70.05\% of the time, compared to only 26.53\% for original distractors. Similarly, easy distractors were classified as the \texttt{Easiest} in 73.17\% of cases, in contrast to 21.21\% in the original dataset. Even without explicit difficulty calibration, more than half of the original distractors fell between the augmented easy and hard ones. The invalid ratio for hard distractors was low at 4.2\%, and no invalid options were found in easy distractors.

\paragraph{Analysis of Two-Way Candidate Generation}

\begin{table}[t]
\begin{adjustbox}{width=0.85\columnwidth,center}
\begin{tabular}{l|cc}
\toprule
\bfseries\makecell[l]{Method} & \bfseries\makecell{Answer\\Generator w/ IR} & \bfseries\makecell{Distractor\\Generator} \\ \midrule
\# of Distractors             & 19.25                                                            & 29.66                                     \\ \midrule
Semantic Diversity            & 0.6928                                                            & 0.6684                                     \\
Semantic Overlap              & \multicolumn{2}{c}{0.2908}                                                                                    \\
Jaccard Overlap               & \multicolumn{2}{c}{0.1281}                                                                                    \\  \bottomrule
\end{tabular}
\end{adjustbox}
\caption{Comparison of distractor statistics between the answer generator with information restriction and distractor generator. Both Jaccard and Semantic Overlap quantify the relationship between the two generation methods.}
\label{tab:detail_stat_generate}
\end{table}

As shown in Table~\ref{tab:detail_stat_generate}, the statistics of the generated distractor candidates highlight the complementary strengths of our two-way candidate generation approach. The Jaccard overlap, which is calculated based on exact matches, is 12.81\%. Moreover, the semantic overlap, which is measured by the average STS score between distractors for each question, is 0.2908. These relatively low values indicate that the answer generator with information restriction and the distractor generator produce distinct and minimally redundant distractor sets.
Moreover, distractors that were generated from the answer generator show slightly greater semantic diversity compared to those from the distractor generator. 
This suggests that applying information restriction on the answer generator increases the semantic variety of the generated distractors and broadens difficulty levels.
Further details on the difficulty distributions are provided in Appendix~\ref{sec:answerVSdistractor}.

\paragraph{Effect of Information Deletion Ratio}
To investigate the impact of information deletion ratios on the quality and diversity of distractors generated with information restriction, we experiment with five deletion ratios: [0.1, 0.2, 0.3, 0.4, 0.5]. For each set of distractors generated from an input question and answer, we evaluated two aspect of semantic diversity and plausibility. 
% Uniqueness refers to the number of distinct distractors generated. 
Semantic diversity is defined as one minus the average pairwise semantic similarity among the generated distractors, while plausibility is measured by the semantic similarity between a distractor and the ground truth answer.

\begin{table}[t] % 데이터 변경 (재 실험했음)
\begin{adjustbox}{width=0.7\columnwidth,center}
\begin{tabular}{c|ccc}
\toprule
\bfseries\makecell{Deletion Ratio} & \bfseries\makecell{Diversity} & \bfseries\makecell{Plausibility} \\ \midrule
0.1                                & 0.6554                        & 0.3404                           \\
0.2                                & 0.6635                        & 0.3189                           \\
0.3                                & 0.6682                        & 0.3066                           \\
0.4                                & 0.6710                        & 0.2978                           \\
0.5                                & 0.6734                        & 0.2920                           \\
\bottomrule
\end{tabular}
\end{adjustbox}
\caption {Average statistics for generated distractors under varying information deletion ratios.}
\label{tab:deletion_ratio_stats}
\end{table}

Table~\ref{tab:deletion_ratio_stats} shows that increasing the information deletion ratio increases semantic diversity of distractors, while their plausibility with respect to the ground-truth answer decreases. This indicates that higher deletion ratios enhance distractor diversity, while lower ratios enhance their plausibility. These findings confirm that deletion ratio adjustments provide a flexible means of controlling distractor difficulty and semantic coverage.

\begin{figure}[t]
\centering
  \includegraphics[width=.8\linewidth]{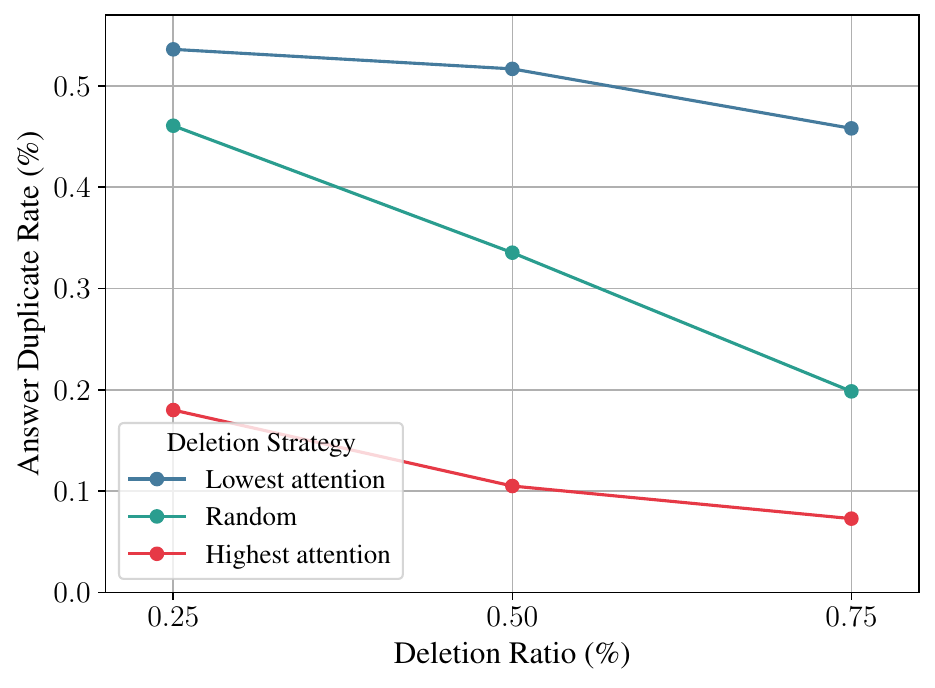}
  \caption {Duplication rate of generated distractors with the correct answers.}
  \label{fig:mask_ans_dup}
  \vspace{-10pt}
\end{figure}

\paragraph{Effectiveness of High-attention Deletion}
To further validate the effectiveness of our highest attention deletion strategy in augmenting distractors, we compared it with two alternative deletion approaches, which are random deletion and lowest attention deletion.
As shown in Figure~\ref{fig:mask_ans_dup}, deleting words with low attention scores or randomly selected words frequently led to high duplication rates of generated distractors matching the ground truth answer, even at relatively high deletion ratios. Specifically, these strategies resulted in more than 40\% duplication rates when 25\% of words were deleted. In contrast, our proposed high-attention deletion approach has significantly low duplication rates, which maintains an answer duplication rate below 20\% at the same deletion ratio. This indicates that selectively deleting high-attention words effectively prevents the answer generator from reproducing the answer, thereby enhancing the efficiency of augmenting plausible distractors.

\subsection{Difficulty-controllable Distractor Generation}

\paragraph{Automatic Evaluation}
To evaluate the difficulty of the generated distractors, we used GPT-4o to assess their relative difficulty. Following the same methodology as the augmented dataset, we presented GPT-4o with cloze passages containing four options: the ground-truth answer, one original distractor, one easy distractor, and one hard distractor generated by our model. Additionally, we compared with those produced by GPT-4o in both 0-shot and 5-shot settings. In the 5-shot setup, we provided examples from the augmented dataset, including eight easy and eight hard distractors for five randomly selected questions. The detailed prompt and few-shot setup used for GPT-4o distractor generation are provided in Appendix~\ref{sec:gpt4o_distractor_prompt}.

\begin{table}[t]
\begin{adjustbox}{width=0.9\columnwidth,center}
\begin{tabular}{lcC{1.3cm}C{1.3cm}C{1.3cm}}
\toprule
\multirow{2}{*}[-3pt]{\textbf{Method}}        & \multirow{2}{*}[-3pt]{\textbf{Type}} & \multicolumn{3}{c}{\textbf{Relative Difficulty}}       \\ \cmidrule{3-5}
                                              &                                      & \textbf{Hardest} & \textbf{Middle}  & \textbf{Easiest} \\ \midrule
\multirowcell{3}[-3pt][l]{GPT-4o (0-shot)}    & Original                                 & 13.64\%          & 35.56\%          & 51.31\%          \\ \cmidrule{2-5}
                                              & Easy                                 & 29.60\%          & 36.36\%          & 33.54\%          \\
                                              & Hard                                 & 56.77\%          & 28.08\%          & 15.15\%          \\ \midrule
\multirowcell{3}[-3pt][l]{GPT-4o (5-shot)}    & Original                                 & 24.35\%          & 38.38\%          & 37.78\%          \\ \cmidrule{2-5}
                                              & Easy                                 & 21.84\%          & 31.06\%          & 46.39\%          \\
                                              & Hard                                 & 53.81\%          & 30.56\%          & 15.83\%          \\ \midrule \midrule
\multirowcell{3}[-3pt][l]{DCDG}               & Original                                 & 20.40\%          & 47.94\%          & 32.36\%          \\ \cmidrule{2-5}
                                              & Easy                                 & 9.25\%           & 32.06\%          & 58.49\%          \\
                                              & Hard                                 & 70.35\%          & 20.00\%          & 9.15\%           \\ \midrule
\multirowcell{3}[-3pt][l]{DCDG\\+ ASDE}        & Original                                 & 19.64\%          & 49.70\%          & 31.66\%          \\ \cmidrule{2-5}
                                              & Easy                                 & 8.22\%           & 30.56\%          & 60.72\%          \\
                                              & Hard                                 & 72.14\%          & 19.74\%          & 7.62\%           \\ \midrule
\multirowcell{3}[-3pt][l]{DCDG\\+ DDDE}        & Original                                 & 22.69\%          & 43.98\%          & 33.73\%          \\ \cmidrule{2-5}
                                              & Easy                                 & 9.34\%           & 32.63\%          & 57.53\%          \\
                                              & Hard                                 & 67.97\%          & 23.39\%          & 8.73\%           \\ \midrule
\multirowcell{3}[-3pt][l]{DCDG\\+ ASDE, DDDE} & Original                                 & 20.04\%          & \textbf{51.80\%} & 28.46\%          \\ \cmidrule{2-5}
                                              & Easy                                 & 6.71\%           & 28.96\%          & \textbf{64.23\%} \\
                                              & Hard                                 & \textbf{73.25\%} & 19.24\%          & 7.31\%           \\
\bottomrule
\end{tabular}
\end{adjustbox}
\caption{Relative difficulty evaluation results of generated distractors. \textit{Original} denotes the distractors in the original CLOTH dataset. }
\label{tab:model_dis_rel_diff}
\end{table}

\begin{table}[t]
\begin{adjustbox}{width=0.9\columnwidth,center}
\begin{tabular}{lcc}
\toprule
\multirow{2}{*}[-2pt]{\textbf{Method}} & \multicolumn{2}{c}{\textbf{Invalid Ratio}} \\ \cmidrule{2-3}
                                       & \textbf{Easy}  & \textbf{Hard}             \\ \midrule
GPT-4o (0-shot)                        & 6.8\%          & 16.9\%                    \\ 
GPT-4o (5-shot)                        & 1.6\%          & 6.8\%                     \\ \midrule
DCDG                                   & 0.9\%          & 6.4\%                     \\ 
DCDG with ASDE                         & \textbf{0.1\%} & 7.0\%                     \\ 
DCDG with DDDE                         & 0.5\%          & 6.3\%                     \\ 
DCDG with ASDE + DDDE                  & 0.2\%          & \textbf{5.1\%}            \\
\bottomrule
\end{tabular}
\end{adjustbox}
\caption{Invalid distractor evaluation results of generated distractors.}
\label{tab:model_dis_invalid}
\vspace{-10pt}
\end{table}

Table~\ref{tab:model_dis_rel_diff} highlights the effectiveness of multitask training in generating distractors with distinct difficulty levels. The model trained with multitask learning outperforms the one fine-tuned only on the main task, showing a greater ability to differentiate between the hardest and easiest distractor categories. In the hard-difficulty setting, 73.25\% of the distractors from the ASDE + DDDE multitask model classify as \texttt{Hardest}, an improvement from 70.35\% in the single task setup. Similarly, for easy distractors, the proportion categorized as \texttt{Easiest} increases from 58.49\% to 64.23\%. On the other hand, GPT-4o struggles with consistency, with only 33.54\% (0-shot) and 46.39\% (5-shot) of easy distractors classified as \texttt{Easiest} For hard distractors, GPT-4o's \texttt{Hardest} rate remains at 56.77\% (0-shot) and 53.81\% (5-shot), which is significantly lower than the proposed approach.

As shown in Table~\ref{tab:model_dis_invalid}, the proposed model with multitask training also achieves notably low invalid ratio of only 0.2\% for easy distractors and 5.1\% for hard distractors. This significantly outperforms GPT-4o, where invalid distractors reach 16.9\% (0-shot hard) and 6.8\% (5-shot hard).

Furthermore, we evaluated the quality of generated distractors by qualitative study and compared our method against established baselines on the original dataset, which confirmed their high quality and consistency. Finally, to further examine the generalizability of our method, we conducted additional experiments using other small LLMs (sLLMs), which showed strong performance in generating high-quality distractors at controlled difficulty levels. For details, see Appendix~\ref{sec:case_study},~\ref{sec:eval_orig_dataset}, and~\ref{sec:various_models} respectively.

\paragraph{Human Evaluation}

\begin{table}[t]
\begin{adjustbox}{width=0.9\columnwidth,center}
\begin{tabular}{lcC{1.2cm}C{1.2cm}C{1.2cm}C{1.2cm}}
\toprule
\multirow{2}{*}[-3pt]{\textbf{Source}} & \multirow{2}{*}[-3pt]{\textbf{Type}} & \multicolumn{3}{c}{\textbf{Relative Difficulty}}       & \bfseries\multirowcell{2}[-3pt]{Invalid\\Ratio} \\ \cmidrule{3-5}
                                       &                                      & \textbf{Hardest} & \textbf{Middle}  & \textbf{Easiest} &                                                 \\ \midrule
\multirowcell{3}[-3pt][l]{Dataset}     & Original                             & 28.8\%           & \textbf{61.6\%}  & 9.6\%            & 3.2\%                                           \\ \cmidrule{2-6}
                                       & Easy                                 & 16.8\%           & 1.6\%            & \textbf{81.6\%}  & \textbf{0.0\%}                                  \\
                                       & Hard                                 & \textbf{54.4\%}  & 36.8\%           & 8.8\%            & 1.6\%                                           \\ \midrule \midrule
\multirowcell{3}[-3pt][l]{Model}       & Original                             & 30.4\%           & \textbf{51.2\%}  & 18.4\%           & 1.6\%                                           \\ \cmidrule{2-6}
                                       & Easy                                 & 24.0\%           & 3.2\%            & \textbf{72.8\%}  & \textbf{0.0\%}                                  \\
                                       & Hard                                 & \textbf{45.6\%}  & 45.6\%           & 8.8\%            & 1.6\%                                           \\
\bottomrule
\end{tabular}
\end{adjustbox}
\caption{Human evaluation results of both our augmented dataset and trained model.}
\label{tab:human_eval}
\end{table}

\begin{table}[t]
\begin{adjustbox}{width=0.67\columnwidth,center}
\begin{tabular}{lc}
\toprule
\textbf{Option Type}    & \bfseries\makecell{Chosen\\Ratio} \\ \midrule
Ground-Truth Answer     & 89.6\%        \\ \midrule
Hard Distractor (Ours)  & 8.6\%         \\
Ground-Truth Distractor & 1.6\%         \\
Easy Answer (Ours)      & 0.2\%         \\
\bottomrule
\end{tabular}
\end{adjustbox}
\caption{Mean ratio of being chosen as the answer for each option type.}
\label{tab:esl_chosen}
\vspace{-10pt}
\end{table}

To assess the student-perceived cognitive load of distractors, we recruited five English as a Second Language (ESL) adult learners. Each participant had less than five years of experience using English. Additional details of human annotation setup is detailed in Appendix~\ref{sec:appendix_human_annotation}.

For evaluation, 50 questions were randomly sampled from the CLOTH test set. Among these, 25 cloze questions were used to evaluate the distractors generated in the augmented dataset, while the remaining 25 questions assessed distractors produced by the multitask-trained model using DCDG with both ASDE and DDDE objectives.
Participants were instructed to solve cloze questions before proceeding with the evaluation. Following the methodology used in our automatic evaluation with GPT-4o, the participants were asked to compare three distractors, which are the ground-truth distractor, an easy distractor, and a hard distractor. The evaluation consisted of two tasks:

\begin{enumerate}
    \item \textbf{Relative Difficulty Assessment:} For each distractor, participants ranked its difficulty as ``\texttt{Hardest},'' ``\texttt{Middle},'' or ``\texttt{Easiest}''.
    \item \textbf{Validity Check:} Participants indicated whether any distractors were more suitable for the blank than the ground-truth answer.
\end{enumerate}

Table~\ref{tab:human_eval} and~\ref{tab:esl_chosen} presents the results of human evaluation on distractor difficulty and validity.
These results align with GPT-4o assessments, confirming that distractors labeled Hard and Easy were consistently perceived as the most and least challenging, respectively. Furthermore, in terms of distractor efficacy, ESL learners selected our Hard distractors more often than ground-truth and Easy distractors. This trend is notably more pronounced compared to native experts (Table~\ref{tab:expert_prob}), validating the model's effectiveness for the target ESL audience.
Additionally, the invalid distractor ratio remained at or below 1.6\%. The results demonstrate that our augmentation method maintains and improves the validity of the original CLOTH dataset.

In addition to evaluations with ESL users, we also conducted a complementary study involving ten proficient English speakers, which confirms pedagogical validity and the consistency of difficulty control across varying user proficiency levels. For details, see Appendix~\ref{sec:human_eval}.

\paragraph{Agreement Between Automatic and Human Evaluation}
To quantify the agreement between automatic and human evaluation, we used the same 50 questions from the human evaluation. For each question, the difficulty rankings from the five ESL annotators were averaged to produce a mean rank per distractor, and Spearman's rank correlation coefficient was computed against GPT-4o's rankings across all questions. The resulting GPT-Human correlation of 0.54 is comparable to the inter-human agreement of 0.62. Given that this task involves ranking only three distractors per question, where even a single-position swap substantially affects the rank correlation, the close alignment between GPT-Human and Human-Human agreement demonstrates that GPT-4o serves as a reliable proxy for human evaluation in this context.

\section{Conclusion}

In this paper, we have proposed a novel method for generating difficulty-controllable distractors, and addressed key challenges in automated assessment. 
Our data augmentation pipeline improved distractor plausibility and diversity through a two-way candidate generation strategy, and ensured quality via a filtering process. By incorporating an advanced normalization method into the difficulty clustering step, we were able to construct a cloze distractor dataset that is annotated with varying difficulty levels.
We further designed a multitask learning framework that enabled the model to generate distractors with controlled difficulty. The resulting model outperformed GPT-4o in both automated and human evaluations, which produced valid distractors that align closely with human-perceived difficulty.
This work contributes to scalable and adaptive assessment in e-learning, supporting systems that tailor distractor difficulty to learner profiles and domain-specific needs.

\section{Limitations}
\paragraph{Consideration of Passage Structure on Difficulty}
While item difficulty is influenced by a wide range of factors, including passage syntax and blank position, this study focuses exclusively on distractor difficulty. We prioritized this dimension as it is a critical factor in item development that is less dependent on learner variability. Future work could enhance the system by incorporating passage readability metrics and syntactic-semantic analyses of blank positions to achieve a more comprehensive difficulty control.

\paragraph{Generalization to Other Question Types}
The information restriction technique is designed for cloze-style questions to assess language proficiency, and its effectiveness for other formats, such as open-ended or math domain questions, are areas that require further study. Similarly, the deletion ratio approach may require additional adaptation to ensure meaningful distractor generation across different structures. Future work could refine deletion strategies or incorporate linguistic constraints to enhance generalizability.

\paragraph{Control Over Continuous Difficulty Levels}
While continuous metrics offer granularity, mapping them directly to pedagogically appropriate levels poses another challenges. Given the lack of prior research and benchmarks for difficulty control, and to avoid arbitrary thresholding, we adopted a binary classification approach grounded in expert calibration. Future work could extend this by leveraging normalized PLM ensemble scores from our difficulty clustering module to define finer-grained, pedagogically calibrated levels.

\section{Ethical Considerations}

Our system is designed as a teacher-centric authoring tool rather than a fully automated test generator. Teachers are responsible for reviewing, revising, and approving all content prior to use. The purpose of the system is not to replace human judgment, but to reduce the time and effort required for educators to create high-quality distractors.

\section*{Acknowledgments}

This research was supported by Culture, Sports and Tourism R\&D Program through the Korea Creative Content Agency grant funded by the Ministry of Culture, Sports and Tourism in 2025 (Project Name: Development of an AI-Based Korean Diagnostic System for Efficient Korean Speaking Learning by Foreigners, Project Number: RS-2025-02413038, Contribution Rate: 45\%); by the IITP (Institute of Information \& Coummunications Technology Planning \& Evaluation) - ITRC (Information Technology Research Center) grant funded by the Korea government (Ministry of Science and ICT) (IITP-2026-RS-2024-00437866, Contribution Rate: 45\%); and by Institute of Information \& communications Technology Planning \& Evaluation (IITP) grant funded by the Korea government (MSIT) (No.RS-2019-II191906, Artificial Intelligence Graduate School Program (POSTECH), Contribution Rate: 10\%).

\bibliography{custom}

\appendix

\section{Differences Between Two Generation Methods}
\label{sec:answerVSdistractor}

This section focuses on the difficulty range of distractors generated by two distinct methods. We assess the difficulty by calculating the QA ensemble system scores for each method. To visualize the difficulty distribution differences, we present a histogram of the score differences between the two generation methods.

\begin{figure}[t]
  \centering
  \includegraphics[width=\linewidth]{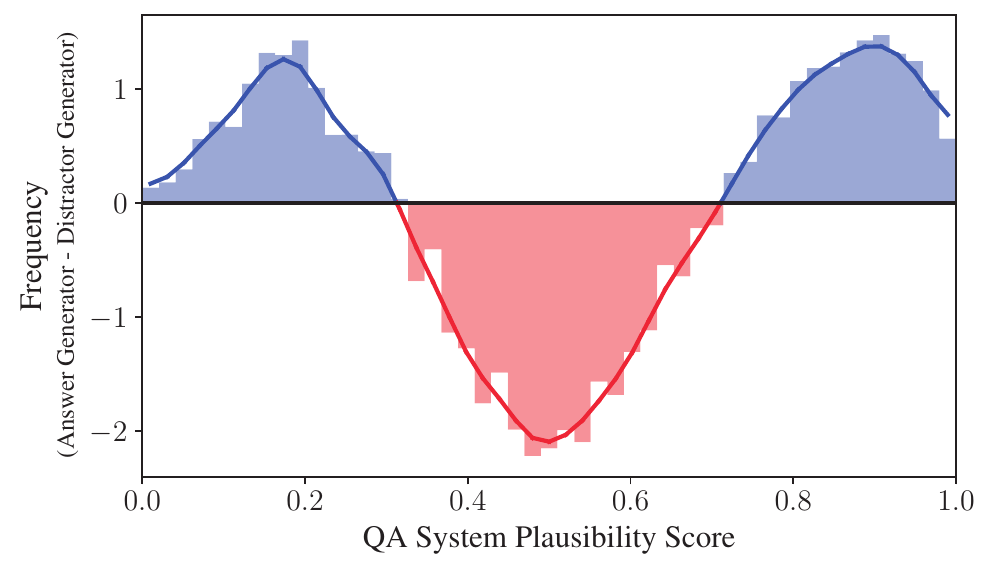}
  \caption {Histogram difference of QA ensemble system scores between two generation methods. Positive values (blue) indicate higher frequencies for the answer generator, while negative values (red) indicate higher frequencies for the distractor generator.}
  \label{fig:generation_mcqa_histogram}
% \vspace{-3pt}
\end{figure}

Figure~\ref{fig:generation_mcqa_histogram} illustrates that the answer generator yields more distractors at both difficulty extremes, indicating a broader difficulty spectrum. This finding suggests that restricting the available information allows the answer generator to generate both easier and more challenging distractors compared to the distractor generator, resulting in a more diverse and balanced set of options. This diversity enhances the robustness of the augmented dataset for training difficulty-controllable distractor generation models.

\section{Pseudo Code of Information Restriction Generation}
\label{sec:info_delete_code}

\begin{algorithm}[t]
\caption{Information Restriction Generation}
\label{alg:info_restrict}
\textbf{Input}: Passage Words $P$, Answer Generator $\mathcal{M}_A$, Deletion Ratios $R$\\
\textbf{Output}: Distractors $\mathcal{D}$
\begin{algorithmic}[1]
\STATE \textbf{Step 1: Sort $P$ with attention scores $S$}
\STATE $A, S \gets \mathcal{M}_A(P)$
\STATE $P_{sorted} \gets sort(P, S, descending)$
\STATE $\mathcal{D} \gets \emptyset$
\FOR{$r \in R$}
    \STATE \textbf{Step 2: Remove key words from passage}
    \STATE $n \gets \lfloor r \cdot |P| \rfloor$
    \STATE $P_r \gets P$
    \FOR{$i = 1$ to $n$}
        \STATE $P_r \gets P_r - P_{sorted}[i]$
    \ENDFOR
    \STATE \textbf{Step 3: Generate distractors}
    \STATE $d_r \gets \mathcal{M}_A(P_r)$
    \STATE $\mathcal{D} \gets \mathcal{D} \cup d_r$
\ENDFOR
\STATE \textbf{Return}: $\mathcal{D}$
\end{algorithmic}
\end{algorithm}

Algorithm~\ref{alg:info_restrict} presents the procedure for generating difficulty-controlled distractors by selectively restricting key information in a given passage. The method begins by identifying important words based on attention scores $S$ assigned by the answer generator $\mathcal{M}_A$. The algorithm ranks the passage words $P$ in descending order of significance according to these attention scores. By iteratively removing a subset of the most influential words at varying deletion ratios $r \in R$, the algorithm creates pruned passages $P_r$ with different levels of missing information. This ensures that the generated distractors are influenced by controlled deletions in contextual cues.

Once the key words are removed, the pruned passage $P_r$ is reprocessed by the answer generator to generate plausible distractors $d_r$. By varying the deletion ratio $r$, the algorithm produces a diverse set of distractors with different levels of semantic plausibility. This controlled variation significantly enhances their suitability for difficulty-tunable multiple-choice questions.

% Algorithm~\ref{alg:info_restrict} presents the procedure for generating difficulty-controlled distractors by selectively restricting key information in a given passage. The method begins by identifying important words based on attention scores assigned by the answer generator. The algorithm ranks passage words in descending order of significance according to these attention scores. By iteratively removing a subset of the most influential words at varying deletion ratios, the algorithm creates pruned passages $P_r$ with different levels of missing information. This ensures that the generated distractors are influenced by controlled deletions in contextual cues.

% Once the key words are removed, the pruned passage $P_r$ is reprocessed by the answer generator to generate plausible distractors $d_r$. By varying the deletion ratio, the algorithm produces a diverse set of distractors with different levels of semantic plausibility, enhancing their suitability for difficulty-controlled multiple-choice questions.

\section{Detailed Information of Ensemble QA System}
\label{sec:qa_system}

\begin{table}[t]
\begin{adjustbox}{width=\columnwidth,center}
\begin{tabular}{lccc}
\toprule
\textbf{Model Name}         & \textbf{Params} & \textbf{LR} & \textbf{Eval Loss} \\ \midrule
albert-xlarge-v2            & 58M             & 3e-06       & 0.6943             \\
albert-xxlarge-v2           & 223M            & 5e-06       & 0.4712             \\
bert-base-uncased           & 110M            & 3e-05       & 1.3764             \\
bert-large-uncased          & 336M            & 1e-05       & 1.1166             \\
conv-bert-base              & 106M            & 3e-05       & 1.2404             \\
deberta-v2-xlarge           & 900M            & 1e-06       & 0.5639             \\
deberta-v2-xxlarge          & 1.5B            & 1e-06       & 0.4986             \\
distilbert-base             & 67M             & 3e-05       & 1.8335             \\
distilroberta-base          & 83M             & 3e-05       & 1.5995             \\
electra-base-discriminator  & 110M            & 1e-05       & 1.3611             \\
electra-large-discriminator & 336M            & 1e-05       & 0.6205             \\
mpnet-base                  & 133M            & 1e-05       & 1.0518             \\
roberta-base                & 125M            & 7e-06       & 1.1444             \\
roberta-large               & 355M            & 3e-06       & 0.9509             \\
spanbert-base-cased         & 110M            & 1e-05       & 1.4093             \\
spanbert-large-cased        & 336M            & 1e-05       & 0.9545             \\
xlnet-base                  & 110M            & 1e-05       & 1.1867             \\
xlnet-large                 & 336M            & 1e-05       & 0.7915             \\
\bottomrule
\end{tabular}
\end{adjustbox}
\caption{Overview of the model candidates used in the ensemble QA system, including the model names, parameter sizes, tuned learning rates (LR), and evaluation loss on the validation set.}
\label{tab:plm_details}
\vspace{-10pt}
\end{table}

To construct a robust and accurate ensemble QA system for estimating distractor difficulty, we evaluated 18 encoder-only models spanning 11 model families with parameter sizes ranging from 67M to 1.5B. Each model was fine-tuned using a batch size of 8 and early stopping, alongside an optimized learning rate to enhance performance and prevent overfitting. ALBERT and DeBERTa models were trained on an NVIDIA RTX6000ADA GPU, while all other models trained on an NVIDIA L40S GPU. Table~\ref{tab:plm_details} presents a detailed overview of the models, including their respective tuned learning rates and evaluation loss scores measured on a validation set.

\section{Prompts for Candidate Filtering}
\label{sec:filtering}

\begin{figure*}[t]
\begin{promptbox}
[SYSTEM]
Your task is to find every possible answer for the cloze question from the candidates provided.
For each candidate, provide a brief explanation about the appropriateness of the candidate.
Then, at the end, **must** provide "Appropriate candidates", with every candidate that could be considered as correct, seperated by a new line.
If there is no appropriate candidate, provide "Appropriate candidates: None".
Evaluate each candidate independently.

[USER]
Masked passage:
<passage with a blank indicated as _____>

Candidates:
<a ground-truth answer and distractor candidates, separated by line break>
\end{promptbox}
\caption{Prompt template used for filtering distractor candidates with GPT-4o mini.}
\label{fig:filtering_prompt}
\vspace{-5pt}
\end{figure*}

The candidate filtering process use a two-shot prompting approach with GPT-4o mini to refine the set of distractor candidates by eliminating those that were more suitable than the ground-truth answer. The detailed prompt structure used for this filtering process is presented in Figure~\ref{fig:filtering_prompt}. This prompt instructed the model to evaluate each candidate independently, provide a justification for its appropriateness, and identify a subset of acceptable candidates. Finally, any distractors included in the list were removed from the candidate pool.

\section{Normalization in QA Ensemble System}
\label{sec:normalization}

\begin{figure}[t]
\centering
\includegraphics[width=\linewidth]{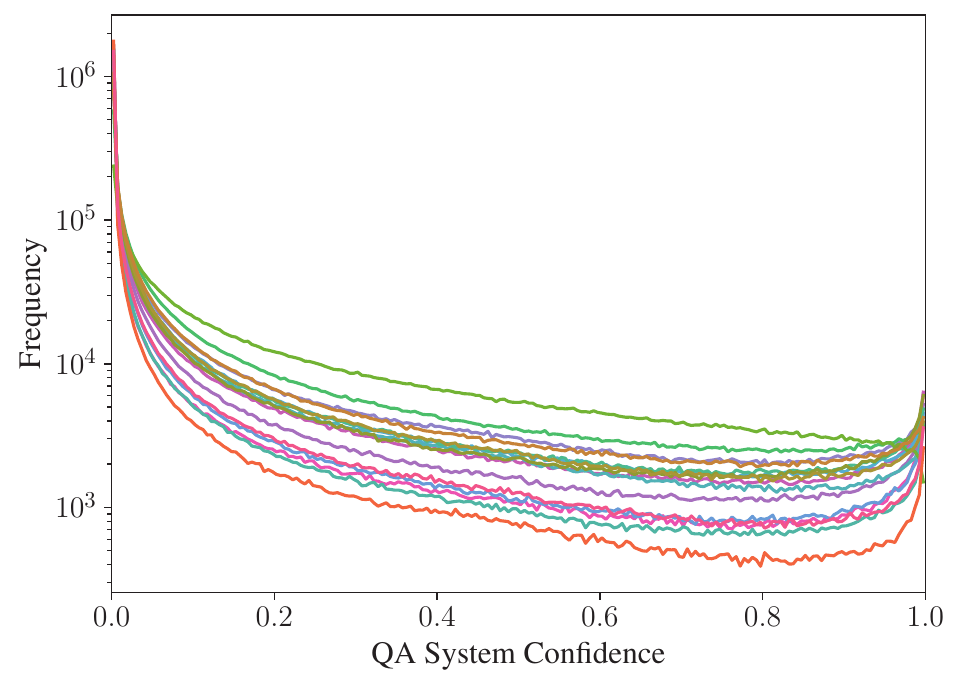}
  \caption {Raw confidence score distribution for each PLM in a log scale.}
  \label{fig:mcqa_distribution}
\vspace{-10pt}
\end{figure}

Normalization of confidence scores across multiple PLMs is crucial to mitigate scale imbalances arising from model-specific and question-dependent variations. Variations in model performance often yield disparate confidence distributions, especially across differing model capacities and question difficulties. To address this, we analyzed the confidence distribution of distractors, which revealed a pronounced right skew. We then compared two normalization techniques, Box-Cox transformation and ranking normalization, to evaluate their effectiveness in stabilizing score distributions.

The confidence score distributions of distractors across PLMs exhibit significant right skewness, as shown in Figure~\ref{fig:mcqa_distribution}, with the y-axis on a log scale. The skewness and variations in model-specific distributions indicate inconsistencies in how PLMs assess distractor plausibility. Such discrepancies can reduce the effectiveness of an ensemble system by introducing aggregation biases.

To address this issue, two normalization methods are considered. Ranking normalization replaces confidence scores with their ranks, effectively mitigating scale imbalance in a simple strategy. However, it only preserves ordinal relationships and disregards absolute score differences. In contrast, Box-Cox normalization transforms right-skewed distributions into a Gaussian form, allowing the retention of relative score magnitudes. To evaluate their effectiveness, we compare the two methods based on their ability to achieve optimal separation in confidence distributions.

The results in Table \ref{tab:qa_ensemble} indicate that Box-Cox normalization achieves a higher degree of separation compared to ranking normalization. Moreover, model combinations with a standard deviation below 0.5\% consistently exhibit higher separation scores under Box-Cox normalization, demonstrating its stability across multiple questions. These findings suggest that Box-Cox normalization more effectively balances score scales, improving the robustness of confidence-based ranking in the QA ensemble system.

\begin{figure*}[ht]
\centering

\begin{adjustbox}{width=\textwidth,center}
\begin{tabular}{lcclcc}
\toprule
\multicolumn{3}{c}{\textbf{Box-Cox Normalization}}                                                                                                                                       & \multicolumn{3}{c}{\textbf{Ranking Normalization}}                                                                                                                                    \\ \cmidrule(r){1-3} \cmidrule(l){4-6}
\textbf{Model List}                                                                                                                                  & $\bm{\mu}$     & $\bm{\sigma}$  & \textbf{Model List}                                                                                                                               & $\bm{\mu}$     & $\bm{\sigma}$   \\ \midrule
\makecell[l]{albert-xxlarge-v2, bert-large, roberta-large,\\xlnet-large}                                                                             & 0.841          & 0.007          & \makecell[l]{albert-xxlarge-v2, bert-large, roberta-large,\\xlnet-large}                                                                          & 0.836          & 0.007          \\
\makecell[l]{albert-xxlarge-v2, conv-bert-base, roberta-large}                                                                                       & 0.840          & 0.006          & \makecell[l]{albert-xxlarge-v2, deberta-v2-xxlarge, mpnet,\\roberta-large, spanbert-base, xlnet-large}                                            & 0.836          & 0.006          \\
\makecell[l]{albert-xxlarge-v2, xlnet-large}                                                                                                         & 0.840          & 0.008          & \makecell[l]{albert-xxlarge-v2, conv-bert-base, roberta-large,\\xlnet-large}                                                                      & 0.836          & 0.007          \\
\makecell[l]{albert-xxlarge-v2, roberta-large, spanbert-base,\\xlnet-large}                                                                          & 0.840          & 0.007          & \makecell[l]{albert-xxlarge-v2, conv-bert-base, deberta-v2-xlarge,\\roberta-large, spanbert-base}                                                 & 0.836          & 0.005          \\
\makecell[l]{albert-xxlarge-v2, roberta-large}                                                                                                       & 0.839          & 0.007          & \makecell[l]{albert-xxlarge-v2, conv-bert-base, roberta-large}                                                                                    & 0.836          & 0.006          \\
\makecell[l]{albert-xxlarge-v2, conv-bert-base, roberta-large,\\xlnet-large}                                                                         & 0.838          & 0.008          & \makecell[l]{albert-xlarge-v2, albert-xxlarge-v2, mpnet,\\roberta-large, spanbert-base, xlnet-large}                                              & 0.835          & 0.009          \\
\makecell[l]{albert-xlarge-v2, albert-xxlarge-v2, conv-bert-base,\\electra-large-dis, roberta-large, xlnet-large}                                    & \textbf{0.837} & \textbf{0.004} & \makecell[l]{albert-xlarge-v2, albert-xxlarge-v2, bert-large,\\roberta-large, spanbert-base, xlnet-large}                                         & 0.834          & 0.012          \\
\makecell[l]{albert-xlarge-v2, albert-xxlarge-v2, bert-large,\\conv-bert-base, deberta-v2-xxlarge, electra-large-dis,\\roberta-large, spanbert-base} & 0.836          & 0.004          & \makecell[l]{albert-xlarge-v2, albert-xxlarge-v2, spanbert-base,\\xlnet-large}                                                                    & 0.834          & 0.007          \\
\makecell[l]{albert-xlarge-v2, albert-xxlarge-v2, bert-large,\\conv-bert-base, deberta-v2-xlarge, electra-large-dis,\\roberta-large, spanbert-base}  & 0.836          & 0.009          & \makecell[l]{albert-xlarge-v2, albert-xxlarge-v2, conv-bert-base,\\deberta-v2-xlarge, roberta-large, spanbert-base,\\xlnet-large}                 & \textbf{0.834} & \textbf{0.004} \\
\makecell[l]{albert-xxlarge-v2, conv-bert-base, roberta-large,\\spanbert-base}                                                                       & 0.836          & 0.005          & \makecell[l]{albert-xxlarge-v2, conv-bert-base, deberta-v2-xxlarge,\\electra-large-dis, roberta-base, roberta-large,\\spanbert-base, xlnet-large} & 0.834          & 0.004          \\
\bottomrule
\end{tabular}
\end{adjustbox}
\captionof{table}{Comparison of the top 10 model combinations by degree of separation under different normalization methods. In the table $\mu$ presents the mean and $\sigma$ presents the standard deviation of performance. For each normalization method, the best-performing model combination with $\sigma < 0.5\%$ is highlighted in bold.}
\label{tab:qa_ensemble}

\begin{promptbox}
[SYSTEM]
Look at the given masked passage and the correct answer, then create 8 distractors (incorrect words) of given difficulty ("Easy" or "Hard") for the cloze question.
At the end, you must write "Easy Distractors:" or "Hard Distractors:", then write down the all the distractors you made, seperated by a new line.

[USER]
Masked passage:
<passage with a blank indicated as _____>

Answer:
<ground-truth answer>

Difficulty:
<'Easy' or 'Hard'>
\end{promptbox}
\captionof{figure}{Prompt template used for GPT-4o distractor generation with difficulty control.}
\label{fig:gpt4o_distractor_prompt}

\vspace{-5pt}
\end{figure*}

\section{GPT-4o Distractor Generation Prompt}
\label{sec:gpt4o_distractor_prompt}

To evaluate difficulty-controllable distractor generation with GPT-4o, we adopted a 5-shot prompting strategy. For each of the five few-shot examples, we provided both `Easy' and `Hard' distractor sets, resulting in a total of ten difficulty-labeled example instances. This design allows the model to learn specific difficulty boundaries by directly contrasting the Easy and Hard examples within the context.

The system prompt which used for GPT-4o distractor generation is in Figure~\ref{fig:gpt4o_distractor_prompt}.

\section{GPT-4o Evaluation Prompts}
\label{sec:gpt_eval}

To ensure the effective evaluation of cloze question distractors, we developed two structured prompt templates designed for GPT-4o. The prompt in Figure~\ref{fig:diff_eval_prompt} assesses the relative difficulty of distractors by ranking them according to how easily they might be confused with the ground truth answer, by providing explanations for why each distractor is incorrect. The second prompt in Figure~\ref{fig:invalid_eval_prompt} detects distractors that are equally or more appropriate than the ground-truth answer, thereby filtering out invalid options that might compromise the overall test's validity.

\begin{figure*}[t]
\begin{promptbox}
[SYSTEM]
Given a cloze question consisting of masked passage, four options and correct answer, evaluate the incorrect options by their **relative** difficulty.
For each incorrect options, provide a brief explanation about their incorrectness.
Then, at the end, you must write "Results:", then write down the all incorrect options in the order of **relative** difficulty.
Start by writing the incorrect answer that is most confusing to distinguish from the correct answer.

[USER]
Masked passage:
<passage with a blank indicated as _____>

Options:
<four options, separated by line break>

The answer is: <ground-truth answer>
\end{promptbox}
\caption{Evaluation prompt template for relative difficulty.}
\label{fig:diff_eval_prompt}
\end{figure*}

\begin{figure*}[t]
\begin{promptbox}
[SYSTEM]
Given a cloze question consisting of masked passage, three options and correct answer, evaluate the option if they are equally suitable or just suitable as the correct answer.
For each options, provide a brief explanation about their incorrectness.
Then, at the end, you must write "Results:", then write down the all options that equally suitable or just suitable as the answer, seperated by a new line.
If there is no options that are equally suitable or just suitable as the answer, provide "Results: None".

[USER]
Masked passage:
<passage with a blank indicated as _____>

Options:
<three distractors, separated by line break>

Answer: <ground-truth answer>
\end{promptbox}
\caption{Evaluation prompt template for assessing invalid distractors.}
\label{fig:invalid_eval_prompt}
\end{figure*}

\section{Consistency with Original Dataset}
\label{sec:eval_orig_dataset}

Since standard multiple-choice questions typically provide only a limited number of distractors per question, categorizing them into difficulty levels results in severe data sparsity. Therefore, while data augmentation is essential to overcome this scarcity, it is still necessary to verify that the generated distractors remain highly consistent with the original dataset. To address this, we evaluated the alignment of the model with both the original and the augmented CLOTH datasets by measuring its ability to generate distractors that replicate the distractor in each dataset.

Specifically, we generated distractors for the test set using the trained model and measured performance via F1 score under an exact match criterion with both original and augmented distractors. Additionally, we compared the model's performance on the original dataset with the state-of-the-art methods~\cite{chiang-2022-cdgp, wang-2023-distractor} to examine the impact of training on the augmented dataset. For the augmented test set, we further analyzed performance across easy and hard difficulty levels to assess the model’s ability to capture difficulty-specific characteristics.

Table~\ref{tab:model_exact_match} shows that our models exhibit a slight decrease in F1 score compared to state-of-the-art on the original dataset, which is reasonable given that our model was trained on more diverse data. On the augmented dataset, our model significantly performs better, achieving F1 scores of 26.93 for easy distractors and 41.61 for hard distractors under the ASDE + DDDE multitask setup. The lower performance on easy distractors likely reflects their broader semantic variability. These results show that the model leverages augmentation to generate controlled distractors while avoiding answers that are misaligned with the original dataset.

\begin{table}[t]
\begin{adjustbox}{width=\columnwidth,center}
\begin{tabular}{cclc}
\toprule
\bfseries\makecell{Eval\\Dataset} & \textbf{Diff}            & \textbf{Method}       & \textbf{F1@10} \\ \midrule
\multirow{5}{*}[-2pt]{Original}   & \multirow{6}{*}[-2pt]{-} & CDGP (2022)           & \textbf{15.37} \\
                                  &                          & Text2Text (2023)      & 14.05          \\ \cmidrule{3-4}
                                  &                          & DCDG                  & 12.84          \\
                                  &                          & DCDG with ASDE        & 12.88          \\ 
                                  &                          & DCDG with DDDE        & 13.23          \\ 
                                  &                          & DCDG with ASDE + DDDE & 13.05          \\ \midrule
\multirow{8}{*}[-2pt]{Augmented} & \multirow{4}{*}{Easy}     & DCDG                  & 26.42          \\
                                  &                          & DCDG with ASDE        & 26.56          \\ 
                                  &                          & DCDG with DDDE        & 25.33          \\ 
                                  &                          & DCDG with ASDE + DDDE & \textbf{26.64} \\ \cmidrule{2-4}
                                  & \multirow{4}{*}{Hard}    & DCDG                  & 41.76          \\
                                  &                          & DCDG with ASDE        & 41.73          \\ 
                                  &                          & DCDG with DDDE        & 41.05          \\ 
                                  &                          & DCDG with ASDE + DDDE & \textbf{41.98} \\ 
\bottomrule
\end{tabular}
\end{adjustbox}
\caption {Distractor exact match results on test set of original and augmented dataset. Diff denotes difficulty.}
\label{tab:model_exact_match}
\end{table}

\section{Experiments with Various sLLMs}
\label{sec:various_models}

\begin{table}[t]
\begin{adjustbox}{width=\columnwidth,center}
\begin{tabular}{lcC{1.3cm}C{1.3cm}C{1.3cm}}
\toprule
\multirow{2}{*}[-3pt]{\textbf{Method}}        & \multirow{2}{*}[-3pt]{\textbf{Type}} & \multicolumn{3}{c}{\textbf{Relative Difficulty}}       \\ \cmidrule{3-5}
                                              &                                      & \textbf{Hardest} & \textbf{Middle}  & \textbf{Easiest} \\ \midrule
\multirowcell{3}[-3pt][l]{GPT-4o (0-shot)}    & Original                             & 13.64\%          & 35.56\%          & 51.31\%          \\ \cmidrule{2-5}
                                              & Easy                                 & 29.60\%          & 36.36\%          & 33.54\%          \\
                                              & Hard                                 & 56.77\%          & 28.08\%          & 15.15\%          \\ \midrule
\multirowcell{3}[-3pt][l]{GPT-4o (5-shot)}    & Original                             & 24.35\%          & 38.38\%          & 37.78\%          \\ \cmidrule{2-5}
                                              & Easy                                 & 21.84\%          & 31.06\%          & 46.39\%          \\
                                              & Hard                                 & 53.81\%          & 30.56\%          & 15.83\%          \\ \midrule \midrule
\multirowcell{3}[-3pt][l]{Gemma 2 9B}         & Original                             & 20.04\%          & \textbf{51.80\%} & 28.46\%          \\ \cmidrule{2-5}
                                              & Easy                                 & 6.71\%           & 28.96\%          & 64.23\%          \\
                                              & Hard                                 & 73.25\%          & 19.24\%          & 7.31\%           \\ \midrule
\multirowcell{3}[-3pt][l]{Llama 3.1 8B}       & Original                             & 22.24\%          & 50.40\%          & 27.76\%          \\ \cmidrule{2-5}
                                              & Easy                                 & 6.81\%           & 27.35\%          & \textbf{65.53\%} \\
                                              & Hard                                 & 70.94\%          & 22.24\%          & 6.71\%           \\ \midrule
\multirowcell{3}[-3pt][l]{Qwen 2.5 7B}        & Original                             & 19.64\%          & 51.10\%          & 29.46\%          \\ \cmidrule{2-5}
                                              & Easy                                 & 6.81\%           & 27.56\%          & 65.33\%          \\
                                              & Hard                                 & \textbf{73.55\%} & 21.34\%          & 5.21\%           \\
\bottomrule
\end{tabular}
\end{adjustbox}
\caption{Relative difficulty evaluation results from various sLLMs. In the table, \textit{Original} denotes the distractors in the original CLOTH dataset.}
\label{tab:various_model_diff}
\end{table}

\begin{table}[t]
\begin{adjustbox}{width=0.6\columnwidth,center}
\begin{tabular}{lcc}
\toprule
\multirow{2}{*}[-2pt]{\textbf{Method}} & \multicolumn{2}{c}{\textbf{Invalid Ratio}} \\ \cmidrule{2-3}
                                       & \textbf{Easy}  & \textbf{Hard}             \\ \midrule
GPT-4o (0-shot)                        & 6.8\%          & 16.9\%                    \\ 
GPT-4o (5-shot)                        & 1.6\%          & 6.8\%                     \\ \midrule
Gemma 2 9B                             & 0.2\%          & 5.1\%                     \\
Llama 3.1 8B                           & 0.2\%          & 5.2\%                     \\
Qwen 2.5 7B                            & \textbf{0.0}\% & \textbf{4.2\%}            \\
\bottomrule
\end{tabular}
\end{adjustbox}
\caption{Invalid distractor evaluation results from various sLLMs.}
\label{tab:various__model_invalid}
\end{table}

To demonstrate the generalizability of our approach, we have conducted additional experiments. We fine-tuned two other sLLMs, Llama 3.1 8B and Qwen 2.5 7B, on our augmented dataset using the DCDG with ASDE + DDDE multitask training strategy. We then evaluated the relative difficulty and invalid ratio of the generated distractors using GPT-4o under the same experimental conditions.

As Table~\ref{tab:various_model_diff} and \ref{tab:various__model_invalid} show, when trained with our proposed method, both Llama 3.1 8B and Qwen 2.5 7B demonstrate a strong capability to generate distractors aligned with specified difficulty levels, while maintaining a low invalid ratio. Although all models perform well, Qwen 2.5 7B shows a slightly better performance than Gemma 2 9B, which confirms the effectiveness and model-agnostic potential of our framework.

\section{Details of Human Annotation}
\label{sec:appendix_human_annotation}

\begin{table}[t]
\centering
\small
\begin{tabular}{lcc}
\toprule
\textbf{Demographic} & \textbf{Experts (N=10)} & \textbf{ESL (N=5)} \\
\midrule
\multicolumn{3}{l}{\textit{Region}} \\
\hspace{3mm}North America & 3 & 3 \\
\hspace{3mm}South America & 5 & 2 \\
\hspace{3mm}Asia & 2 & - \\
\midrule
\multicolumn{3}{l}{\textit{Gender}} \\
\hspace{3mm}Male & 8 & 4 \\
\hspace{3mm}Female & 2 & 1 \\
\bottomrule
\end{tabular}
\caption{Demographic statistics of human annotators.}
\label{tab:annotator_demographics}
\end{table}

For all human evaluations conducted in this study, we recruited participants through the Amazon Mechanical Turk platform. To ensure ethical standards, participants were explicitly informed that the collected data would be used exclusively for research purposes before beginning the tasks. We compensated participants with a fixed hourly rate in accordance with the legal minimum wage standards of the authors' nationality.

Participants were categorized into two groups based on their English proficiency duration:
\begin{itemize}
    \item \textbf{Experts:} Users with over 10 years of English usage experience.
    \item \textbf{ESL Learners:} Users with less than 5 years of English usage experience.
\end{itemize}

The demographic breakdown of the recruited participants is detailed in Table \ref{tab:annotator_demographics}.

\section{Evaluation with English Expert}
\label{sec:human_eval}

\begin{table}[t]
\begin{adjustbox}{width=0.67\columnwidth,center}
\begin{tabular}{lcc}
\toprule
\textbf{Option Type}    & \bfseries\makecell{Chosen\\Ratio}  \\ \midrule
Ground-Truth Answer     & 83.2\%                            \\ \midrule
Hard Distractor (Ours)  & 9.4\%                             \\
Ground-Truth Distractor & 7.0\%                             \\
Easy Answer (Ours)      & 0.4\%                             \\
\bottomrule
\end{tabular}
\end{adjustbox}
\caption{Mean ratio of being chosen as the answer by expert English user.}
\label{tab:expert_prob}
\end{table}

\begin{table}[t]
\begin{adjustbox}{width=0.67\columnwidth,center}
\begin{tabular}{lc}
\toprule
\textbf{Option Type}    & \bfseries\makecell{Invalid\\Ratio} \\ \midrule
Hard Distractor (Ours)  & 8.4\% \\
Ground-Truth Distractor & 5.0\% \\
Easy Answer (Ours)      & 1.0\% \\
\bottomrule
\end{tabular}
\end{adjustbox}
\caption{Invalid ratio for each distractor type by expert English user.}
\label{tab:expert_invalid}
\end{table}

In this section, we present our experiment results from \textbf{expert} English users. We had recruited 10 proficient English users following Appendix~\ref{sec:appendix_human_annotation} and had them evaluate 50 questions by choosing the most suitable answer from four options, which are the ground-truth answer, the ground-truth distractor, our easy distractor, and our hard distractor.

Table~\ref{tab:expert_prob} and \ref{tab:expert_invalid} demonstrate that our model's hard distractors are sophisticated enough to be chosen over the original dataset's distractors even by proficient users. The near-zero selection rate for easy distractors confirms that our difficulty control is effective across different user proficiency levels. Additionally, while our hard distractors are challenging than the original distractors, their invalid ratio remains low.

\section{Qualitative Analysis}
\label{sec:case_study}

In this section, we present the results of two selected questions from the CLOTH test set. For each question, we provide the passage, the ground-truth answer, three original distractors, eight augmented distractors, and eight generated distractors on both easy and hard difficulty levels.

\begin{table*}[t] 
\begin{adjustbox}{width=0.95\textwidth,center}
\begin{tabular}{cC{2.3cm}C{2.3cm}C{2.3cm}C{2.3cm}C{2.3cm}}
\toprule
\textbf{Passage}     &  \multicolumn{5}{m{13cm}}{Yesterday I was tidying up my room. I found an old box of my father's. He gave it to me two years ago. It was fascinating to discover some of my father's childhood photos. He once told me that he wrote to people all over the world, and they sent him letters, too. As a result, he had a book of interesting stamps. People also gave him things from different countries, such as a silk from Japan, a little doll from England, and a small model ship from Australia. My father even kept he tickets from his first football match! It made me think about looking after my collection of \_\_\_\_\_ pictures books and magazines.} \\ \midrule
\textbf{Answer}      & \multicolumn{5}{C{12.8cm}}{different} \\ \midrule
\multirowcell{10}[-2pt]{\textbf{Distractors}} & \multirowcell{2}[-2pt]{\textbf{Original}} & \multicolumn{2}{c}{\textbf{Augmented}} & \multicolumn{2}{c}{\textbf{Generated}} \\ \cmidrule(r){3-4} \cmidrule(l){5-6}
                                              &                                           & \textbf{Easy}      & \textbf{Hard}     & \textbf{Easy}      & \textbf{Hard}     \\ \cmidrule{2-6}
                                              & new                                       & disappointing      & old               & difficult          & old               \\
                                              & old                                       & difficult          & interesting       & easy               & beautiful         \\
                                              & interesting                               & disgusting         & travel            & funny              & interesting       \\
                                              &                                           & bad                & colorful          & expensive          & new               \\
                                              &                                           & cheap              & school            & small              & other             \\
                                              &                                           & comic              & many              & common             & various           \\
                                              &                                           & boring             & stamps            & same               & good              \\
                                              &                                           & funny              & useful            & bad                & similar           \\
\bottomrule
\end{tabular}
\end{adjustbox}
\caption{An example of augmented and generated distractors.}
\label{tab:case_1}
\vspace{-10pt}
\end{table*}

Table~\ref{tab:case_1} demonstrates an example from the CLOTH-M subset, which is derived from middle school English examinations. It can be observed that both the augmented distractors and generated distractors contain a diverse range of distractors across easy and hard difficulty levels. The augmented easy distractors primarily consist of negative words that contrast with the passage’s main theme of diverse travel experiences. In contrast, the augmented hard distractors include words that are more consistent with the passage’s content. The model-generated easy distractors show a diverse semantic range but still include words that do not align well with the passage. Meanwhile, the hard distractors by the model exhibit a stronger semantic connection to the passage than the easy distractors.

\begin{table*}[t]
\begin{adjustbox}{width=0.95\textwidth,center}
\begin{tabular}{cC{2.3cm}C{2.3cm}C{2.3cm}C{2.3cm}C{2.3cm}}
\toprule
\textbf{Passage}     &  \multicolumn{5}{m{13cm}}{When I was a boy, every holiday that I had seemed wonderful. My parents took me by train or by car to a hotel by the sea. All day, I seem to remember, I played on the sands with strange excited children. We made houses and gardens, and watched the tide destroy them. When the tide went out, we climbed over the rocks and looked down at the fish in the rock-pools. In those days the sun seemed to shine always brightly and the water was always warm. Sometimes we left the beach and walked in the country, exploring ruined houses and dark woods and climbing trees. There were sweets in one's pockets or good places where one could buy ice-creams. Each day seemed a life-time. Although I am now thirty-five years old, my idea of a good \_\_\_\_\_ is much the same as it was. I still like the sun and warm sand and the sound of waves beating the rocks. I no longer wish to build any sand house or sand garden, and I dislike sweets. However, I love the sea and often feel sand running through my fingers. Sometimes I wonder what my ideal holiday will be like when I am old. All I want to do then, perhaps, will be to lie in bed, reading books about children who make houses and gardens with sands, who watch the incoming tide, who make themselves sick on too many ices.} \\ \midrule
\textbf{Answer}      & \multicolumn{5}{C{12.8cm}}{holiday} \\ \midrule
\multirowcell{10}[-2pt]{\textbf{Distractors}} & \multirowcell{2}[-2pt]{\textbf{Original}} & \multicolumn{2}{c}{\textbf{Augmented}} & \multicolumn{2}{c}{\textbf{Generated}} \\ \cmidrule(r){3-4} \cmidrule(l){5-6}
                                              &                                           & \textbf{Easy}      & \textbf{Hard}     & \textbf{Easy}      & \textbf{Hard}     \\ \cmidrule{2-6}
                                              & house                                     & house              & day               & party              & day               \\
                                              & garden                                    & impression         & summer            & festival           & time              \\
                                              & tide                                      & job                & time              & week               & trip              \\
                                              &                                           & month              & fun               & job                & weekend           \\
                                              &                                           & game               & experience        & work               & life              \\
                                              &                                           & event              & weekend           & dream              & week              \\
                                              &                                           & evening            & pleasure          & meal               & place             \\
                                              &                                           & night              & life              & dinner             & vacation          \\
\bottomrule
\end{tabular}
\end{adjustbox}
\caption{An example of augmented and generated distractors.}
\label{tab:case_2}
\vspace{-10pt}
\end{table*}

Table~\ref{tab:case_2} presents a longer passage from the CLOTH-H subset, which is based on high school English examinations. Similar to the previous case, a clear distinction in difficulty levels can be observed among the distractors. The augmented hard distractors contain words related to the passage's theme of holiday experiences, whereas the augmented easy distractors, though grammatically correct, are semantically unrelated to the main topic. A similar pattern emerges in the generated distractors, where the hard distractors include words associated with holidays and time, while the easy distractors mostly contain less relevant terms.

\end{document}